\documentclass[runningheads]{llncs}
\usepackage[T1]{fontenc}
\usepackage{xcolor}
\usepackage{graphicx,verbatim}
\usepackage[colorlinks=true]{hyperref}
\usepackage{amsmath, amssymb}
\usepackage{stmaryrd}
\usepackage{dsfont}
\usepackage{multicol}
\usepackage{blindtext}
\usepackage{graphics}
\usepackage{caption}
\usepackage{subcaption} 
\usepackage{multirow}  
\usepackage{arydshln}  
\usepackage{booktabs}
\usepackage{cleveref}

\begin{document}
\title{Robust sensitivity control in digital pathology via tile score distribution matching}
\titlerunning{TSM}
%
\author{Arthur Pignet\inst{1, \dagger} \and
John Klein\inst{1} \and
Geneviève Robin\inst{1, *} \and
Antoine Olivier\inst{1, *, \dagger}
}
\index{Pignet, Arthur}
\index{Klein, John}
\index{Robin, Geneviève}
\index{Olivier, Antoine}
\authorrunning{A. Pignet et al.}
%
\institute{$^1$ Owkin, Inc\\
{$\dagger$ Corresponding author.} \\
{$^*$ Equal senior authorship.} \\
{\texttt{\{arthur.pignet, john.klein, genevieve.robin,
antoine.olivier\}@owkin.com}}
}

\maketitle              
\begin{abstract}

Deploying digital pathology models across medical centers is challenging due to distribution shifts. Recent advances in domain generalization improve model transferability in terms of aggregated performance measured by the Area Under Curve (AUC). However, clinical regulations often require to control the transferability of other metrics, such as prescribed sensitivity levels. We introduce a novel approach to control the sensitivity of whole slide image (WSI) classification models, based on optimal transport and Multiple Instance Learning (MIL). Validated across multiple cohorts and tasks, our method enables robust sensitivity control with only a handful of calibration samples, providing a practical solution for reliable deployment of computational pathology systems.


\end{abstract}

\section{Introduction}
\label{sec:intro}
Deep learning (DL) models are used in Computational Pathology (CPath) to analyze whole slide images (WSI) in a variety of medical contexts~\cite{liu_artificial_2019,krithiga_breast_2021,saillard_validation_2023}. However, their deployment in the clinic is limited by their ability to generalize well beyond the training context, impaired by inherent data variability due to the use of different scanners, staining and labeling protocols~\cite{stacke_measuring_2021}.
To overcome this, recent works in DL for CPath have applied domain generalization (DG) techniques, designed to increase the robustness of predictive models to distribution shifts between training and evaluation~\cite{tellez_quantifying_2019,chen_synthetic_2021,gulrajani2021in,jarkman_generalization_2022,greenspan_text-guided_2023,chen_towards_2024,filiot_phikon-v2_2024}. 

\begin{figure}[t]
    \centering
    \includegraphics[width=0.45\linewidth]{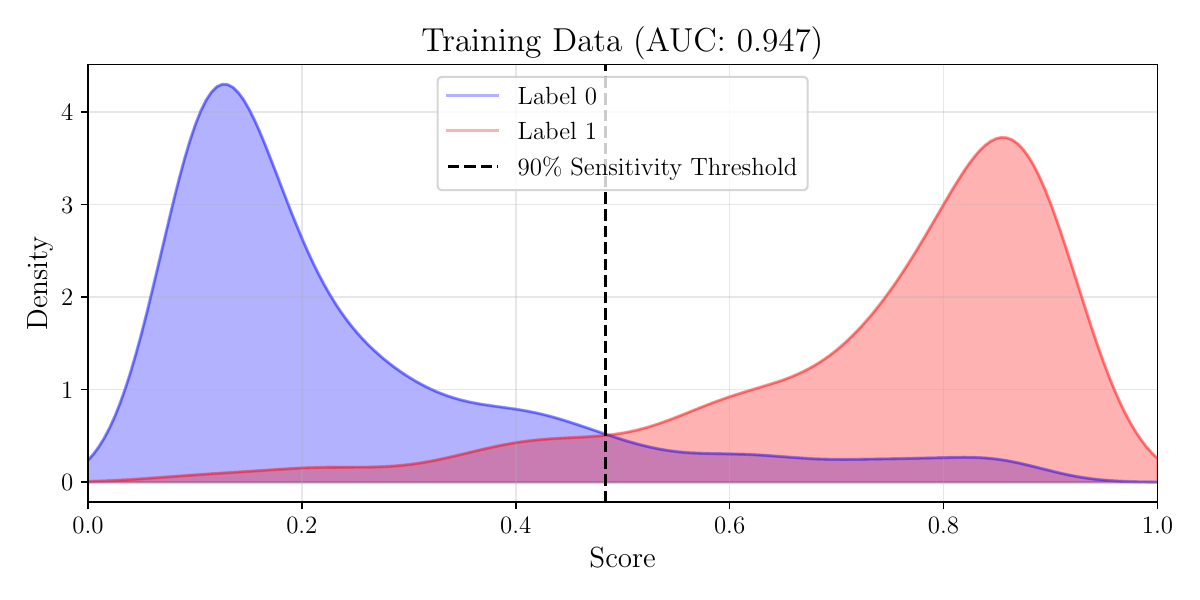}
    \includegraphics[width=0.45\linewidth]{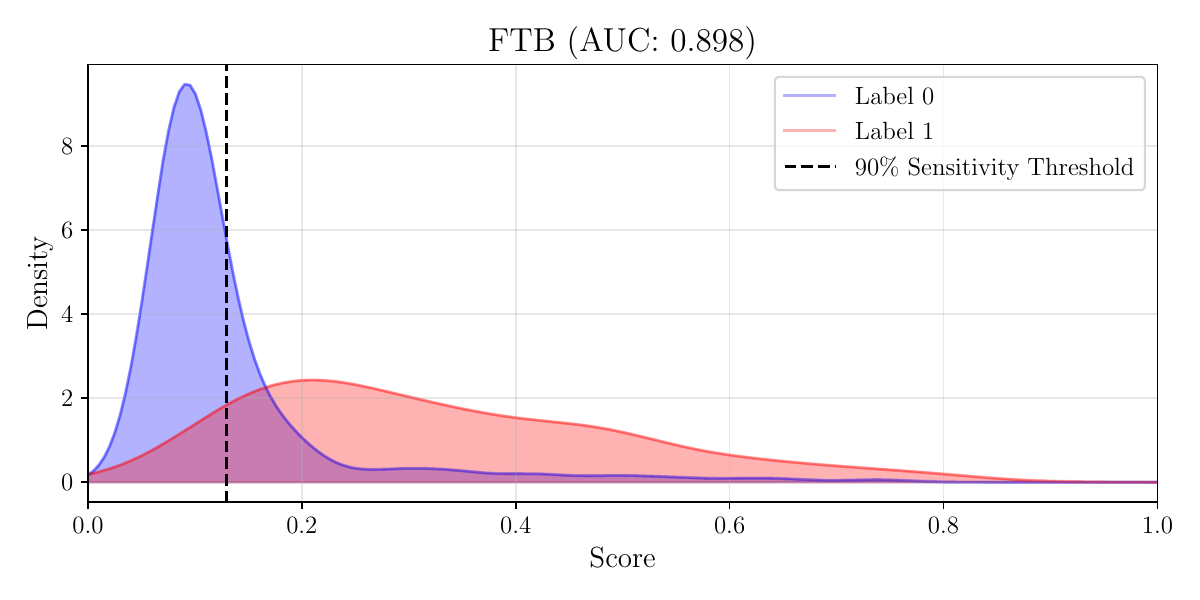}
    \caption{Distribution of a DL model's prediction scores on the training (top) and validation cohorts (FTB, bottom) with comparable AUC (train: 94.7\%, val: 89.8\%). Dashed line correspond to the threshold associated to a sensitivity of 90.0\%, (train: 0.45, FTB: 015).}
    \label{fig:sensitivity-thresholds}
\end{figure}

The performance of DG methods is usually evaluated with the Area Under the ROC Curve (AUC), which may fail to capture their generalization capacity in clinical contexts, where sensitivity and specificity are also crucial~\cite{kleppe_area_2022}. 
Indeed, the sensitivity/specificity trade-off is often controlled via a threshold transforming continuous scores into binary labels. Even when models preserve the AUC, the distribution of predicted scores may vary between cohorts~\cite{roschewitz_automatic_2023}, impairing the threshold's capacity to yield consistent sensitivity and specificity~\cite{echle_artificial_2022}, as illustrated in~\Cref{fig:sensitivity-thresholds}. To solve this issue, existing works have relied on calibration procedures to adjust score distributions.

Strictly speaking, probabilistic calibration refers to the transformation of prediction scores into actual class membership probabilities~\cite{dawid_well-calibrated_1982}, for instance using temperature scaling~\cite{DBLP:conf/icml/GuoPSW17}. While probabilistic model calibration is a way to control sensitivity, it is a stronger requirement aiming to control sensitivity at \emph{any level}. In practice, a weaker requirement is that models achieve \emph{a fixed sensitivity level} prescribed by the clinical context (say, $90\%$). Thus, aligned with existing work, we use `calibration' in its more general meaning, referring to a model's capacity to output similar scores at training and inference time. In practice, this is often achieved using calibration data from the deployment center, to adjust the model's threshold after training~\cite{saillard_validation_2023,roschewitz_automatic_2023}.

In this paper, we develop a new methodology called Tile-Score Matching (TSM) to control the sensitivity in WSI binary classification problems. We focus on Multiple Instance Learning (MIL) models based on the Chowder architecture~\cite{courtiol2018classification}, which has been applied to many WSI classification problems~\cite{saillard_validation_2023,xu_weak_2020,xu_deep_2021,courtiol_deep_2019}. In Chowder, each WSI is divided into tissue patches called tiles; prediction scores are computed at the tile-level, then aggregated into a single WSI-level score. We leverage this property by working at the tile level to calibrate the distribution of prediction scores. 
TSM is most similar to Unsupervised Prediction Alignment (UPA)~\cite{roschewitz_automatic_2023}, which also calibrates the distribution of prediction scores. However, while UPA matches score distributions at the WSI level, TSM operates at the tile level, increasing the number of available calibration samples by several orders of magnitude. As a result, while UPA requires hundreds of WSI for calibration, TSM requires less than 30. 

Our contributions are presented as follows. In \Cref{sec:method}, we introduce TSM, a new threshold calibration method which matches the distribution of tile-level prediction scores to a reference distribution using optimal transport (OT), and accounts for prevalence shift with importance sampling (IS). We also provide theoretical evidence of sensitivity control in particular cases. In \Cref{sec:experiments}, we demonstrate empirically that TSM controls the sensitivity across several indications and classification tasks\footnote{Our code, model and features are available at \url{https://github.com/owkin/tsm.git}.}. Our experiments also show that, contrary to existing works, TSM controls sensitivity in extremely low data and prevalence regimes, when only 5 positive samples are available for calibration.

\section{Tile-Score Distribution Matching}
\label{sec:method}
\begin{figure}[t]
    \centering
    \includegraphics[width=\linewidth]{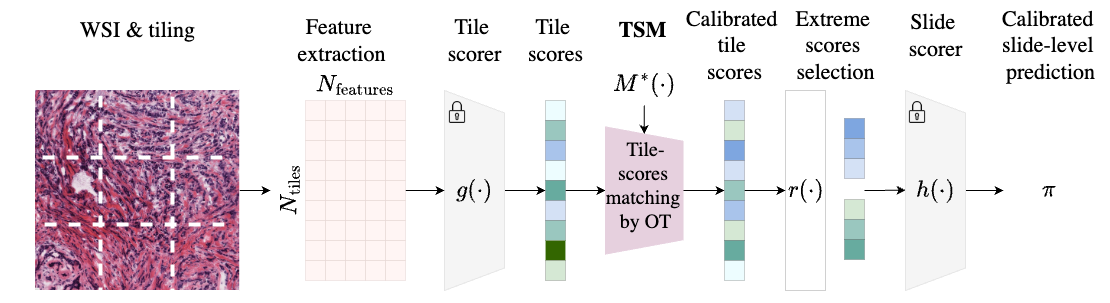}
    \caption{Graphical overview of the method.}%
    \label{fig:TSM-illustration}
\end{figure}

In this section, we present our method and explain how it can be combined with Chowder~\cite{courtiol2018classification} to target pre-specified sensitivity levels, as illustrated in \Cref{fig:TSM-illustration}. For clarity, we restrict to sensitivity, but the extension of the method to reach pre-specified specificity levels is straightforward by simply considering negative samples instead of positive samples.

\subsubsection{General framework.} Let $\mathbb{S}$ denote the observation space of WSI, $S^t$ a WSI from the reference cohort (possibly used during training) and $Y^t$ the associated binary label. Respectively, denote $S^c$ a WSI from the calibration population and $Y^c$ the associated binary label. In what follows, we use the superscript $\{t,c\}$ to denote the membership to the training or calibration cohorts. Let $(\mathbb{S}\times\{0,1\},\mathcal{S}\otimes\mathcal{P}\{0,1\},\mu)$ be a measurable space, and assume that  $(S^t,Y^t)$ and $(S^c,Y^c)$ admit positive density functions with respect to the base measure $\mu$.
%

Similarly to other MIL models, Chowder identifies a WSI $S\in\mathbb{S}$ to a collection of tiles belonging to $S$. For simplicity, we assume that each WSI has the same number of tiles $N$, and denote $S = (T_1,\ldots,T_N)$, $T_i\in\mathbb{T} := \mathbb{R}^{d \times d}$ for all $1\leq i \leq N$ (where $d \times d$ is the tile size). By design, the prediction function $f$ is the composition of a tile-level scoring function $g: \mathbb{T}\mapsto\mathbb{R}$, a selection  function $r: \mathbb{R}^N\mapsto\mathbb{R}^{2k}$, which ranks and selects the top-$k$ and bottom-$k$ tile-scores, and a predictor $h:\mathbb{R}^{2k}\mapsto\mathbb{R}$:
\begin{equation}
f(S) = h(r(g(T_1), \ldots, g(T_N))).
\end{equation}
Finally, the predicted patient label is given by $\mathds{1}\left\lbrace f(S) \geq \tau\right\rbrace$. Usually, $\tau$ is adjusted post-training to achieve
the prescribed sensitivity level $\sigma$,
\begin{equation}
    \mathbb{P}_{S, Y}\left(f(S)>\tau | Y=1\right) = \sigma,
\end{equation}
by leveraging held-out calibration data (see, \emph{e.g.}, the methodology deployed in~\cite{saillard_validation_2023}). As previously highlighted, $\tau$ often does not transfer to external cohorts, in the sense that the achieved level of sensitivity will drift apart from the prescribed target $\sigma$ due to distributional shifts. For instance, in~\Cref{fig:sensitivity-thresholds}, $\tau = 0.45$ is calibrated during training to achieve $\sigma_{\text{train}} = 0.9$, but yields $\sigma_{\text{val}} = 0.32$ on the validation cohort.
We develop the TSM methodology to mitigate this lack of transferability, by matching the distribution of tile scores $g(T_i)$ between the training and application domains. 
\subsubsection{Tile-score distribution matching.}
By construction, for $1\leq i\leq N$ and for $k\in\{t,c\}$, $T_i^k$ admits a positive density function on $\mathbb{T}$. For simplicity of exposition, we assume that, in each cohort, tiles are i.i.d. conditionally to the slide label. Thus, for $1\leq i,j\leq N$, $k\in\{t,c\}$ and for $l\in\{0,1\}$, $\mathbb{P}(T_i^k | Y^k = l) = \mathbb{P}(T_j^k | Y^k = l)$. 
This assumption is required to prove the theoretical control of sensitivity, yet, it may not hold in practice, as adjacent tissue samples have correlated spatial structure. We note however that the sparsity of the tiling can be controlled, for instance by extracting only a subset of tiles, serving as an effective way to reduce the spatial correlation between the considered tiles. Besides, we show experimentally in~\cref{sec:experiments} that sensitivity is controlled in real-life scenarios.

As the tile score function $g$ is continuous, the random variables $(X_i := g(T_i^k))_{1 \leq i \leq N}$ are also i.i.d. conditionally to the label $Y^k$. Denoting $\omega^k = \mathbb{P}(Y^k=1)$ the prevalence in cohort $k$, the density function of a tile score $X$ is given for $k \in \{c,t\}$ by
\begin{equation}
    \rho_X^k =\omega^k \rho^k_{X | Y=1} + (1-\omega^k) \rho^k_{X | Y=0}.
\end{equation}

We now explain how the tile score distribution $\rho_X^c$ is matched, up to an adjustment w.r.t. prevalence, to the reference distribution $\rho_X^t$. The Monge formulation of OT writes, for two measures $a$ and $b$ on $\mathbb{R}$:
\begin{equation}
\label{eq:transport-map}
M^* = \underset{M \in \mathcal{M}_{a \rightarrow b}}{\arg\min}  \int_{\mathbb{R}} |M(x) - x|^2 \text{d}a(x),
\end{equation}
where $\mathcal{M}_{a \rightarrow b}$ is the set of Borel measurable functions such that $M_{\#}a = b$, \emph{i.e.}, $b$ is the push-forward measure of $a$ through $M$.
Since $X$ is one-dimensional,~\eqref{eq:transport-map} has a closed-form solution, which is the monotonous map obtained through quantile matching~\cite{mccann_exact_1999}. The optimal map is given by
$M^* = F_{b} \circ {F_{a}}^{-1},$
where $F_{a}$ and $F_{b}$ are the cumulative distribution functions of measures $a$ and $b$ and ${F_{a}}^{-1}$ is the generalized inverse of $F_{a}$.
TSM consists in applying~\eqref{eq:transport-map} to $a = \rho_X^c$ and 
\begin{equation}
b = \omega^c\rho^t_{X|Y=1} + (1-\omega^c)\rho^t_{X|Y=0},
\end{equation}
a reweighted version of $\rho^t_X$ which accounts for prevalence shift between training and validation. We now prove that TSM controls the sensitivity level when $\omega^c=1$, \textit{i.e.}, when the calibration set contains only positive labels.
\begin{theorem}
\label{thm:sensitivity-control}
    Let $\tau\in\mathbb{R}$, and denote by $\text{sens}_{train}(\tau)$ and $\text{sens}_{val}(\tau)$ the sensitivities associated to threshold $\tau$ on the training and validation cohorts. Assume that the calibration set contains only positive examples, \emph{\emph{i.e.}}, $\omega^c=1$. Then, 
    \begin{equation}
    \text{sens}_{val}(\tau) = \text{sens}_{train}(\tau).
    \end{equation}
\end{theorem}
\begin{proof}
After calibration, the sensitivity on the validation cohort is given by
$\text{sens}_{val}(\tau) = \mathbb{P}_{M^*\#(\rho^c_X)}(f(S^c) > \tau | Y=1).$
By construction of the transport map $M^{*}$, and with $\omega^c=1$, we have $M^*_{\#}(\rho^c_X) = \rho^t_{X | Y=1}$ and $\rho^c_{X|Y=1} = \rho^c_X$. Thus,
\begin{align*}\text{sens}_{val}(\tau) &= \mathbb{P}_{M^*_{\#}(\rho^c_X)}(f(S^c) > \tau | Y=1) \\
&= \int_{f^{-1}(]\tau, 1])}{M^*_{\#}(\rho^c_{X})(x_1) \dots M^*_{\#}(\rho^c_{X})(x_N) dx_1 \dots dx_N} \\
&= \int_{f^{-1}(]\tau, 1])}{\rho^t_{X | Y=1}(x_1) \dots \rho^t_{X | Y=1}(x_N) dx_1 \dots dx_N} \\
&= \mathbb{P}(S^t \in f^{-1}(]\tau, 1]) | Y=1) = \mathbb{P}(f(S^t) > \tau | Y=1)=\text{sens}_{train}(\tau).
\end{align*}
\end{proof}
\Cref{thm:sensitivity-control} implies that TSM, applied to a calibration set drawn conditionally on $Y=1$, effectively ensures transferability of the model's sensitivity. However, as shown in~\cref{sec:experiments}, using a calibration set containing both positive and negative samples can lead to similar sensitivity transfers.

\begin{lemma}
\label{lemma:rank}
    Let $M^*_d: R^{d} \mapsto R^{d}$ be the $d$ multi-dimensional (component-wise) application of $M^*$ and $r$ the ranking function of the Chowder model. Then, 
    \begin{equation}
       M^*_N\circ r = r \circ M^*_{2k}.
    \end{equation}
\end{lemma}
\begin{proof}
    The function $r$ in Chowder is a ranking function~\cite{courtiol_deep_2019}. Thus, the monotonicity of $M^*$ concludes the proof.
\end{proof}
\Cref{lemma:rank} states that the set of $2k$ tiles selected by the ranking layer of the Chowder model is invariant through calibration by TSM. This means in particular that interpretation properties are preserved, as the regions of the WSI used by the final prediction function $h$ are unchanged.
\Cref{lemma:rank} also has computational implications, as the Monge map only needs to be applied after the ranking layer $r$, reducing the number of operations from $N$ to $2k$.

In practice, the densities $\rho^c_X$ and $\rho^t_X$ are unknown, and we approximate them by the weighted sum of Dirac masses centered on each data point, \emph{\emph{i.e.}}, 
\begin{equation}
\widehat{\rho^c_X} = \frac{1}{n_c} \sum_{i=1}^{n_c} \delta_{x^c_{(i)}}, \quad \text{and } \quad
 \widehat{\rho^t_X} = \frac{1}{n_t} \sum_{j=1}^{n_t} \delta_{x^t_{(j)}},
\end{equation}
where $( x^c_{(i)} )_{1\leq i\leq n_c}$ and $ (x^t_{(j)})_{1\leq j\leq n_t}$ are the empirical tile scores obtained by applying $g$ to the training and validation set of tiles, respectively. Consequently, $F_{ \widehat{\rho^c_X}}$ and $F_{ \widehat{\rho^t_X}}$ are staircase functions. 
To provide more flexibility in $M^*$, we resort to a linear interpolation between function jumps. 

\section{Experiments}
\label{sec:experiments}
\subsubsection{ER, PR and HER2 status prediction in breast cancer.} In a first study, we predict the status of estrogen receptor (ER), progesterone receptor (PR) and human epidermal growth factor receptor 2 status (HER2) in breast cancer. We use TCGA-BRCA as training cohort. For ER and PR, we use the 1076 labeled slides. The prevalence of ER+ (resp. PR+) patients in the training dataset is $78\%$ (resp. $68\%$). For the HER2 task, we use the reliable labels from ~\cite{tcga_her2_2012}, yielding 801 slides ($15\%$ HER2+). Two external datasets are used for validation. For the 3 endpoints, we use the BCNB dataset~\cite{xu_predicting_2021} with early breast cancer core-needle biopsies collected from 1,058 patients, along with ER, PR and HER2 statuses ($79\%$ ER+, $75\%$ PR+, $26\%$ HER2+). Additionally, for HER2 status prediction, we use the Herohe dataset~\cite{jimaging8080213} ($41\%$ HER2+).

\subsubsection{MSI status prediction in colorectal cancer.}
In a second study, we predict high Microstatellite instability (MSI-H) against low instability or stability (MSI-L or MSS). The training set is the combination of TCGA-COAD, TCGA-READ, and a private dataset from Medipath laboratories (France). The prevalence of MSI-H in the training dataset is 18\%. Five external datasets are used for validation. Cypath is a private collection of 698 H\&E and H\&E\&S biopsies from 698 patients ($36\%$ MSI-H) digitized in France; Cypath-HE and Cypath-HES cohorts are obtained by splitting Cypath to account for the variations in staining conditions. Neogenomics is a second private collection of 198 biopsies and 200 resections ($43\%$ overall); it also further splits into Neogenomics-resections and Neogenomics-biopsies. FTB is a private collection of 602 patients ($26\%$ MSI-H).

\subsubsection{Training setup.}
For the 4 prediction tasks, a pre-trained feature extractor is first used to extract low-dimensional representations for tiles. We use Phikon, for the 3 breast-related tasks and a feature extractor tailored for colorectal cancer for MSI status prediction, as in~\cite{saillard_validation_2023}. Chowder models are then trained to predict a slide-level score with repeated cross-validation (3 repeats and 5 splits). For each split, 5 Chowder models are trained, resulting in an ensemble of 75 models. During validation, TSM is applied to each model independently, and the resulting 75 calibrated models are ensembled to produce the final prediction. 

\subsubsection{TSM controls sensitivity and preserves ROC curves.}
\begin{figure}[t]
\centering
\begin{subfigure}{0.4\textwidth}
    \includegraphics[width=\textwidth]{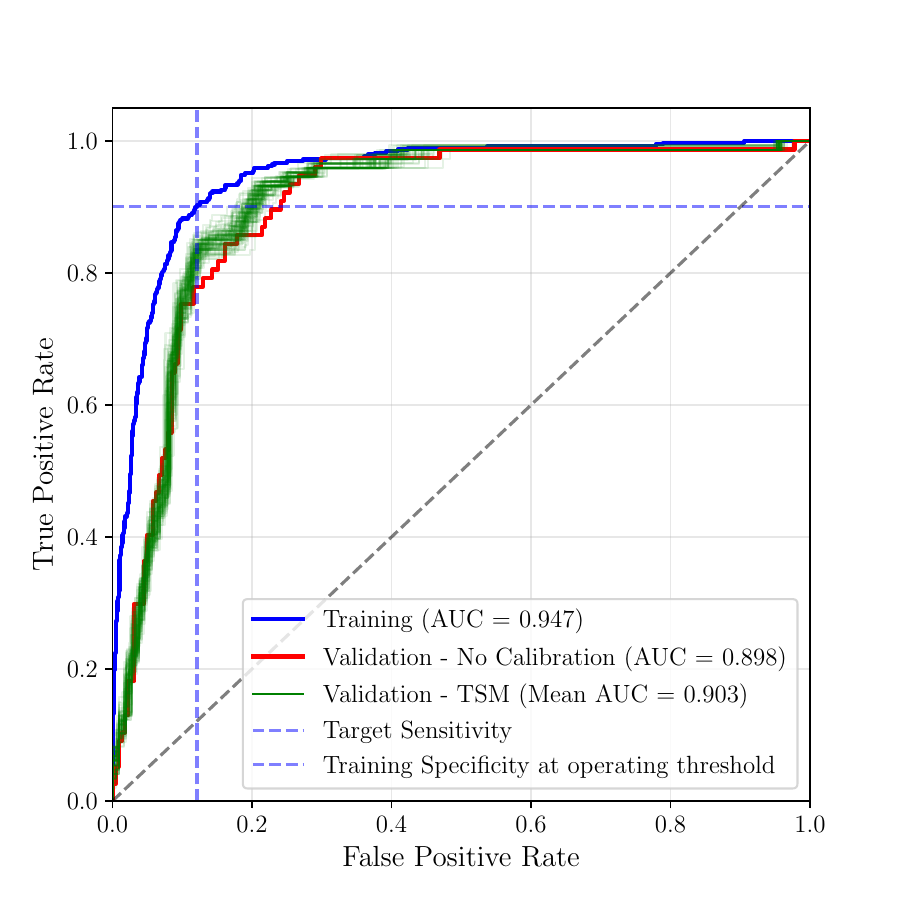}
    \caption{ROC curves.}
    \label{fig:roc}
\end{subfigure}
\begin{subfigure}{0.4\textwidth}
    \includegraphics[width=\textwidth]{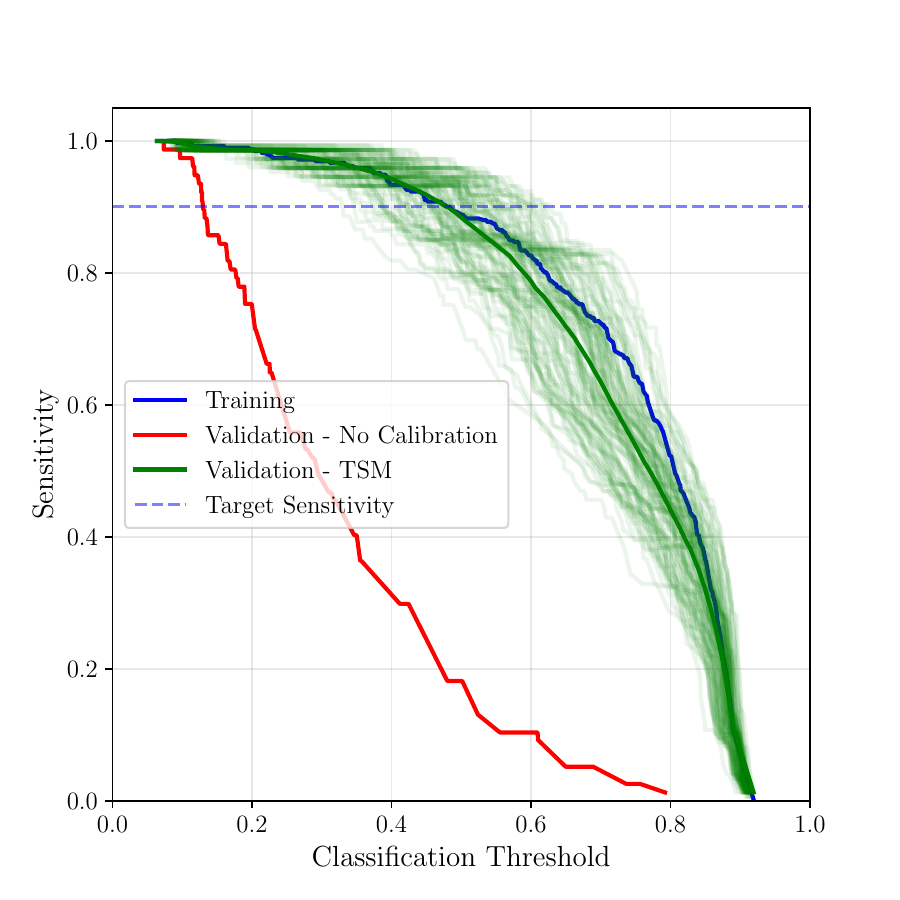}
    \caption{Sensitivity vs thresholds.}
    \label{fig:sens}
\end{subfigure}
\label{fig:results}
\caption{Metrics computed on both training and FTB validation cohorts for the MSI classification task, with and without TSM. While the validation ROC curves maintain their distinct profiles, the calibrated sensitivity curves converge towards the training curve, suggesting effective calibration. Each shaded green curve represents an individual sampling of the calibration set.}
\end{figure}

\Cref{fig:roc} depicts the ROC curves with (in green) and without (in red) TSM calibration, and shows that TSM does not impact the ability of the model to rank patients correctly; \Cref{tab:model-performance} also highlights that the AUC is preserved by TSM. \Cref{fig:sens} illustrates the good transfer of the sensitivity/threshold curve from the training cohort (in blue) to the calibrated external validation cohort (in green). On the contrary, the sensitivity/threshold curve without calibration (in red) shows that applying the training threshold to the external cohort without calibration would lead to a dramatic drop of sensitivity (from 90\% to 20\%). 

\begin{table}[htbp]
\caption{Transfer of model AUC across different cohorts. The training AUC is estimated via cross-validation; the validation AUC is computed on external cohorts, with and without TSM calibration.}
\label{tab:model-performance}
\centering
\begin{tabular}{lccccccccc}
\toprule
\textbf{Task}& ER & PR & HER2 & HER2 & \multicolumn{5}{c}{MSI Status}\\
\midrule
\textbf{Ext. Cohort} & \multicolumn{3}{c}{BCNB} & HEROHE & \multicolumn{2}{l}{Cypath} & \multicolumn{2}{l}{Neogenomics} & FTB\\
\midrule
 &\multicolumn{4}{c}{} & HE & HES & Bio. & Res. & \\
\midrule
\textbf{Train} & 0.904 & 0.809 & 0.818 & 0.818 & \multicolumn{5}{c}{0.947}\\
\textbf{Validation} & 0.818 & 0.816& 0.643 & 0.569 & 0.917 & 0.899 & 0.947 & 0.977 & 0.898\\
\textbf{Calibrated Val.} & 0.825 & 0.818 & 0.633 & 0.604 & 0.913 & 0.898 & 0.958 & 0.977 & 0.903\\
\bottomrule

\end{tabular}
\end{table}

\subsubsection{Comparison with prior art.}
\begin{figure}[t]
    \centering
    \includegraphics[width=\linewidth]{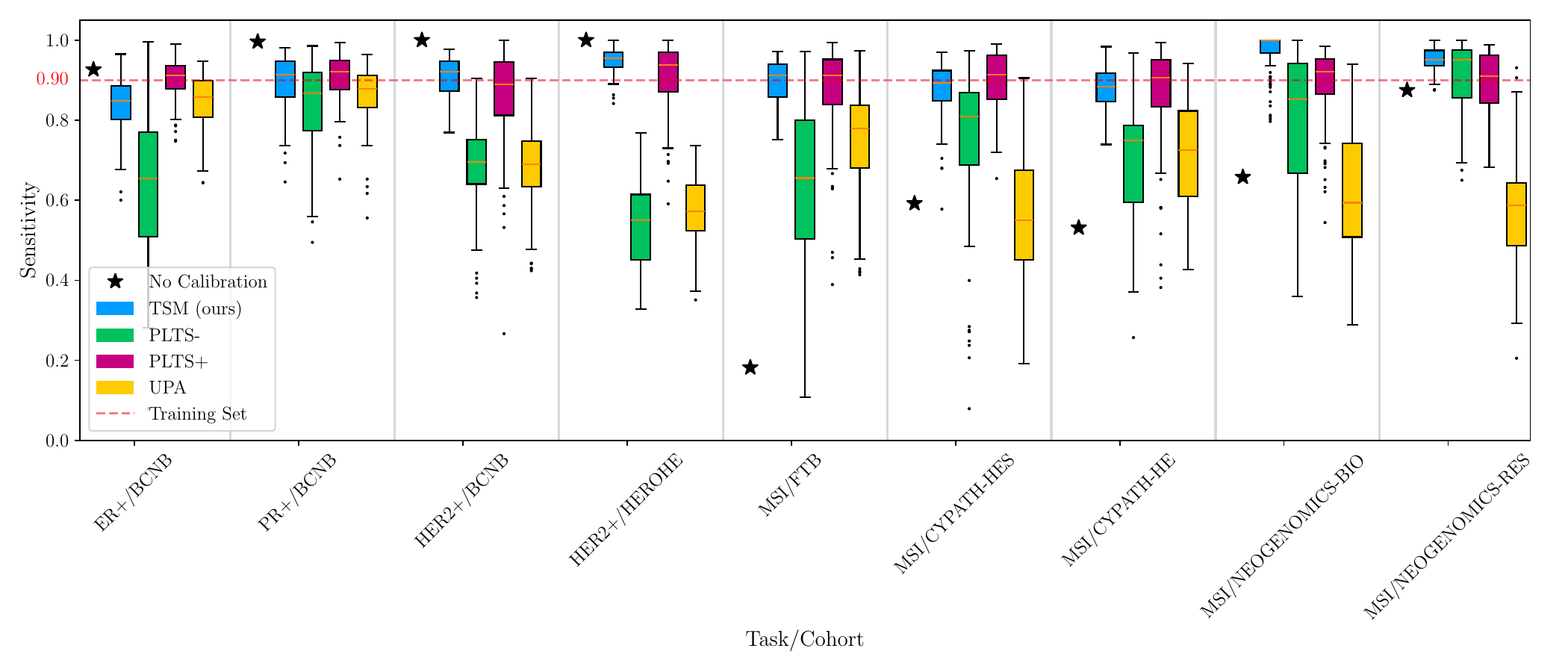}
    \caption{Sensitivity achieved by TSM, UPA, and PLTS+/- on several tasks and validation cohorts using 30 slides for calibration, across 100 experiments.}%
    \label{fig:sensitivity-all}
\end{figure}

We compare TSM to two sensitivity control methods from the literature: UPA~\cite{roschewitz_automatic_2023}, discussed in \Cref{sec:intro}, and Patient level threshold selection (PLTS, see~\cite{saillard_validation_2023}). PLTS leverages a calibration set composed of $m$ WSIs with \textit{constant labels}, denoted $\left( S_i^c\right)_{i=1}^m$, and matches the threshold $\tau$ to a quantile of the scores computed on the calibration data. In the case of sensitivity control, and if $\pi^{c}_{i} = f (S_i^{c})$ denote the ordered calibration scores, \emph{\emph{i.e.}}, $\pi^{c}_{1}\leq .. \leq \pi^{c}_{m}$, , it sets $\tau = \pi_i^c$ where the integer $i$ is set to the largest such that $sens\left(\pi_i^c\right) \geq \sigma$. This methodology can be used with positive samples to control sensitivity (PLTS+) or with negative samples to control specificity (PLTS$-$). In~\Cref{fig:sensitivity-all}, we report the sensitivities obtained by all methodologies using $30$ WSI for calibration. We observe that TSM and PLTS+ reach the desired sensitivity while keeping relatively close to the target, while UPA falls short in this low data regime. TSM and PLTS+ have similar average sensitivity, but TSM exhibits less variability. 
\subsubsection{TSM outperforms PLTS in low prevalence regime.}
\begin{figure}[h!]
    \centering
    \includegraphics[width=0.7\linewidth]{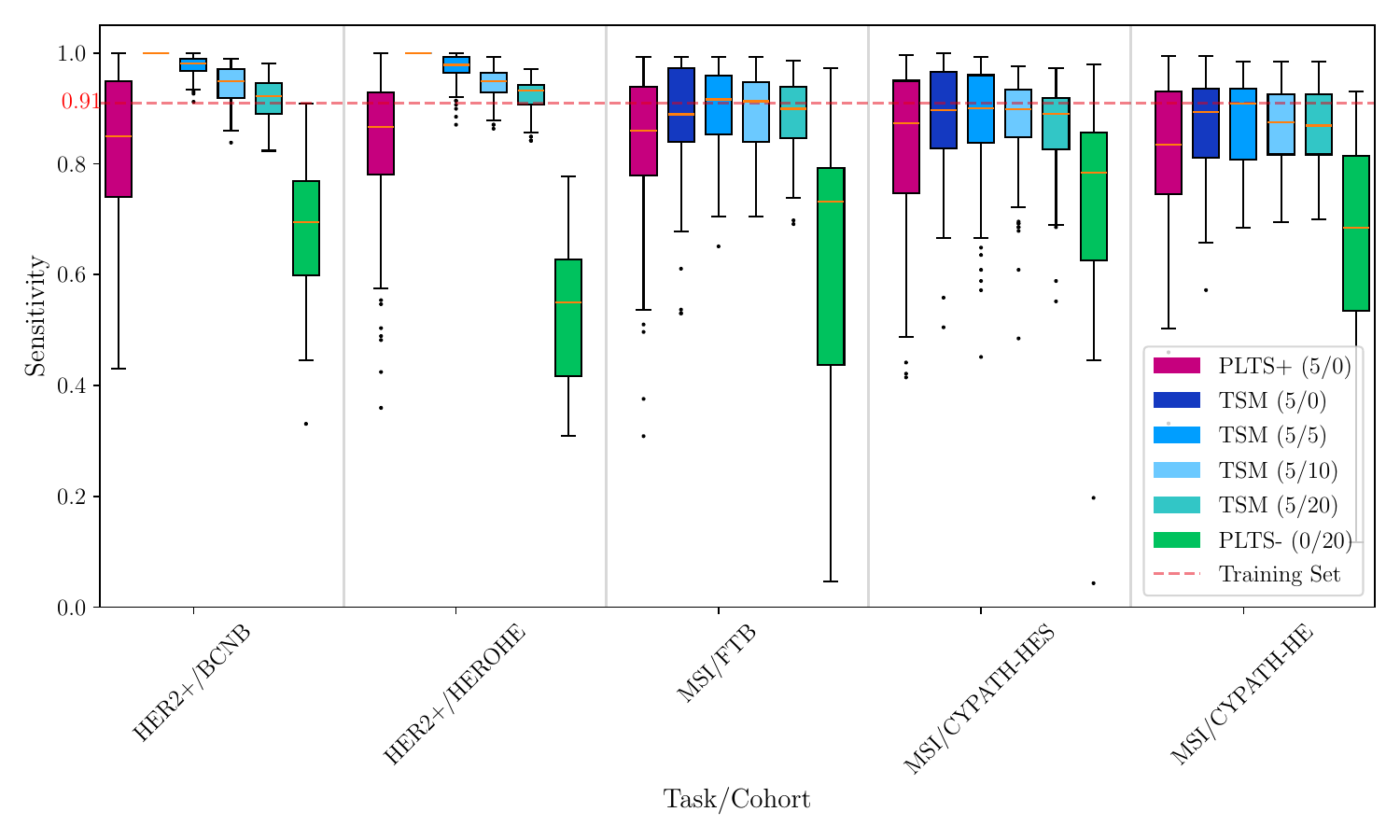}
    \caption{Sensitivity achieved by PLTS+ and TSM using 5 positive samples for calibration. For TSM, we add up to 20 additional negative samples; the ratio (positive/negative) is provided in the legend.}%
    \label{fig:sensitivity-all-5}
\end{figure}

In~\Cref{fig:sensitivity-all-5}, we further compare TSM and PLTS+ in the more challenging setting where only a handful of positive samples can be used for calibration. In this experiment, we focus on the low prevalence tasks with less than 30\% positive samples, and PLTS+ and TSM are allowed only 5 positive samples for calibration. By design, PLTS+ can only use positive samples; on the contrary, TSM is able to also leverage the negative samples. Enriched with negative samples, TSM~(5/20) improves over PLTS+, both in terms of targeted sensitivity and variability. 

\section{Conclusion}
In this paper, we propose a novel threshold calibration method to control sensitivity in WSI binary classification tasks. Using only a handful of calibration samples, TSM outperforms existing works in terms of sensitivity control on several indications and classification tasks. The main shortcoming of TSM is its specificity to the Chowder architecture which, although frequently used for WSI analysis, remains limiting. Thus, future work includes the extension of TSM to other MIL modeling approaches~\cite{gadermayr_multiple_2024}. Another direction of future research is the analysis of TSM in settings where the exact prevalence of positive samples is unknown. For instance, in the context of the deployment of TSM in a new hospital, the historical prevalence of the center can be used, as well as the indication prevalence from existing epidemiological data or public health statistics.  

    

\begin{credits}
\subsubsection{\ackname} The results presented here are in part based upon data generated by Medipath, Cypath and the TCGA Research Network: \url{https://www.cancer.gov/tcga.} Authors would like to thank Alexandre Filiot and Auriane Riou for their valuable help and insights to conduct the downstream evaluations, and Lucas Fidon for his feedback.

\subsubsection{\discintname}
Persons affiliated with Owkin own stock-options in the company (A.P., J.K., G.R., A.O.).
\end{credits}

%
%
%

%

\bibliographystyle{splncs04}
\bibliography{paper-2982.bib}

\end{document}